\documentclass[letterpaper, 10 pt, conference]{ieee_format/ieeeconf}  

\IEEEoverridecommandlockouts                              

\usepackage[noadjust]{cite}
\usepackage{threeparttable}
\usepackage{url}
\usepackage{xcolor}
\usepackage{kotex}

\usepackage{graphicx}
\usepackage{amssymb} 
\usepackage{amsmath}

\usepackage{siunitx}
\usepackage{float}
\usepackage{dsfont}
\usepackage{multicol}
\usepackage{bm}
\usepackage{booktabs}
\usepackage{tabularx}
\let\labelindent\relax
\usepackage{enumitem}

\usepackage{multirow}

\usepackage{caption}
\title{\LARGE \bf
    Enhancing Navigation Efficiency of Quadruped Robots

    via Leveraging Personal Transportation Platforms
}

\author{Minsung Yoon\textsuperscript{1} and Sung-Eui Yoon\textsuperscript{1}\textsuperscript{\textdagger}
\thanks{
    \textsuperscript{1}M. Yoon and S. Yoon are with the School of Computing at the Korea Advanced Institute of Science and Technology (KAIST), Daejeon, 34141, Republic of Korea. E-mails: {\tt\small minsung.yoon@kaist.ac.kr, sungeui@kaist.edu}.
    {\textsuperscript{\textdagger}}S. Yoon is a corresponding author.
    }}

\begin{document}
    \maketitle
    \thispagestyle{empty}
    \pagestyle{empty}

    \begin{abstract}
Quadruped robots face limitations in long-range navigation efficiency due to their reliance on legs. To ameliorate the limitations, we introduce a Reinforcement Learning-based Active Transporter Riding method (\textit{RL-ATR}), inspired by humans' utilization of personal transporters, including Segways. The \textit{RL-ATR} features a transporter riding policy and two state estimators. The policy devises adequate maneuvering strategies according to transporter-specific control dynamics, while the estimators resolve sensor ambiguities in non-inertial frames by inferring unobservable robot and transporter states. Comprehensive evaluations in simulation validate proficient command tracking abilities across various transporter-robot models and reduced energy consumption compared to legged locomotion. Moreover, we conduct ablation studies to quantify individual component contributions within the \textit{RL-ATR}. This riding ability could broaden the locomotion modalities of quadruped robots, potentially expanding the operational range and efficiency.

\end{abstract}
    \section{Introduction}

Quadruped robots have proven remarkable versatility in a range of applications, from space and nature exploration to surveillance and rescue missions~\cite{arm2023scientific, lindqvist2022multimodality, delmerico2019current, bellicoso2018advances}. 
Recent research has enhanced their locomotion capabilities over challenging terrains, including rough, slippery, deformable, and moving surfaces~\cite{lu2023whole, fankhauser2018robust, Suyoung2023deformter, 8772165, da2021learning}. 
Nevertheless, their four-legged designs inherently encounter limitations in speed and energy efficiency during long-range tasks, coupled with the risk of mechanical failures due to cumulative stress from repetitive foot contacts.

To alleviate these limitations, researchers have developed multi-modal locomotion systems integrating wheels or skates into legs, enabling both walking and driving~\cite{bjelonic2021whole, bjelonic2020rolling, lee2024learning, bjelonic2022offline, jelavic2023lstp, zhou2023max, geilinger2020computational, bjelonic2018skating, chen2023locomotion}. 
These systems enhance navigation speed and energy efficiency on specific surfaces such as ice or paved roads.
However, these permanently attached devices can increase hardware costs of each quadruped robot and compromise navigation efficiency in each modality due to cumbersome leg designs~\cite{tenreiro2006overview, valsecchi2023towards}.

Meanwhile, humans augment their mobility using shared transportation platforms, such as Segways and hoverboards, as needed~\cite{nguyen2004segway, siddhardha2019quadrotor, zapata2024flyboard, hendohover, rosrobosavvybalance2017, hover1hoverboards}.
These platforms allow users to traverse large areas quickly with minor physical exertion required for control and balance.   
Moreover, these platforms are shareable among users, regardless of kinematics and size variations.

Inspired by these advantages, recent studies on humanoid robots have developed platform-maneuvering controllers by adapting standing controllers that adjust the Center of Mass (CoM) or foot angles to modulate platform inclinations~\cite{xin2017torque, gong2019feedback, kimura2018riding, rajendran2022towards, chen2019feedback}.
However, these conventional model-based approaches often constrain the platform's mobility due to modeling inaccuracies, uncertainties, and conservative constraints. 
Moreover, they exhibit limited resilience to unexpected situations, such as momentary foot contact loss due to external perturbations.
To mitigate these limitations, we employ a model-free Reinforcement Learning (RL) approach to develop adaptive and resilient control strategies, enhancing robustness against environmental disturbances and domain variations.

\begin{figure}[t]
    \vspace{0.2cm}
    \centering 
    \includegraphics[width=\columnwidth]{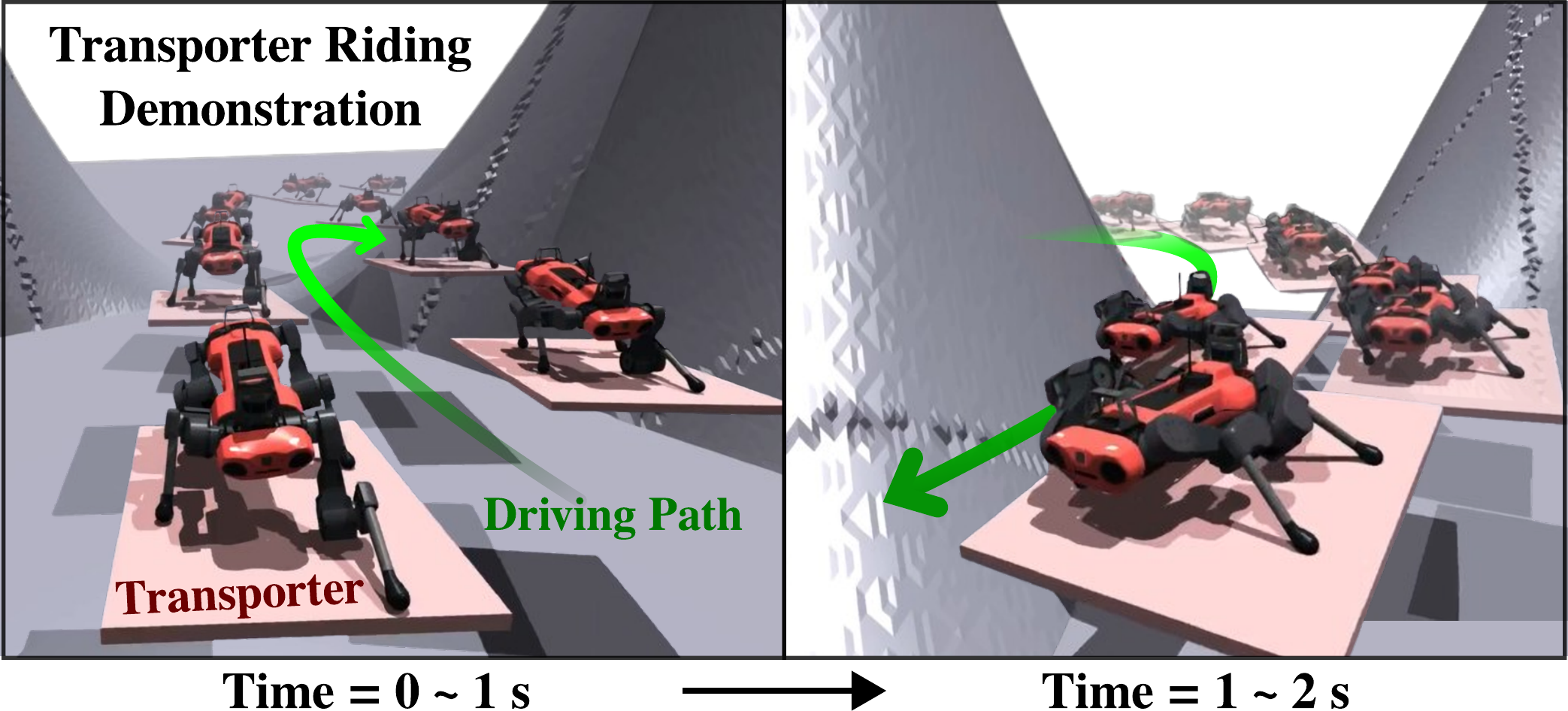}
    \vspace{-0.5cm}
    \caption{
        Demonstration of the \textit{RL-ATR}: Quadruped robots utilizing personal transportation platforms (transporters) with adept riding ability for efficient long-range navigation. Specific transporter dynamics are detailed in Sec.~\ref{sec:tp_types}.
    }
    \label{fig:main_fig}
    \vspace{-0.45cm}
\end{figure}

Therefore, we aim to ensure that quadruped robots adeptly utilize transportation platforms, also known as transporters, for efficient long-range navigation, as shown in Fig.~\ref{fig:main_fig}. 
To the best of our knowledge, we believe this work is the first effort to incorporate active transporter riding skills into quadruped robots, facilitating multi-modal locomotion with riding capabilities.
To achieve this, robots need to adeptly maneuver transporters according to specific platform dynamics while maintaining stability on moving platforms.
This necessitates understanding inertia effects, as described by Newton's Laws of Motion, and counteracting the fictitious inertial forces that arise from acceleration changes of the underlying platform. 

\textbf{Main Contributions}. 
We introduce a Reinforcement Learning-based Active Transporter Riding method (\textit{RL-ATR}), a low-level quadrupedal controller maneuvering transporters to satisfy velocity commands through transporter's motions.
To develop these adept riding skills using RL, we construct simulation environments incorporating quadruped robots and transporters with specific dynamics detailed in Sec.~\ref{sec:tp_types}. 

The \textit{RL-ATR} features an active transporter riding policy and two state estimators, optimized through RL and system identification.
The policy modulates quadrupedal postures to induce adequate platform tilts for transporter control, while preserving stability.
These estimators enhance the situational awareness of the policy in non-inertial frames by estimating privileged states, like underlying platform's movements, and intrinsic domain parameters from historical sensor data.

Furthermore, we adopt a grid adaptive curriculum learning approach~\cite{margolis2024rapid} to effectively cover command spaces. 
This is crucial for effective policy learning, enabling the policy to progressively confront and master challenging situations.

To validate the effectiveness of the \textit{RL-ATR} in simulation, \\
we evaluate command tracking accuracy across various transporter and robot models, encompassing A1, Go1, Anymal-C, and Spot robots~\cite{taheri2023study}.
In addition, we measure the mechanical Cost of Transport (CoT)~\cite{bjelonic2018skating} to validate the energy efficiency of utilizing transporters for long-range navigation, compared to legged locomotion.
Lastly, we conduct ablation studies to analyze the contributions of components within the \textit{RL-ATR}.

    \section{Variable Notation}
For clarity, we present variable notations used throughout this manuscript.
In Cartesian space, $\boldsymbol{p}$, $\boldsymbol{v}$, and $\boldsymbol{\dot{v}} \in \mathbb{R}^3$ denote position, velocity, and acceleration, respectively.
$\boldsymbol{\theta}$, $\boldsymbol{\omega}$, $\boldsymbol{\alpha}$, and $\boldsymbol{\tau} \in \mathbb{R}^3$ indicate Euler angles using the XYZ convention, angular velocity, angular acceleration, and torque, respectively.
For precise specification of physical quantities, we utilize superscripts to denote reference coordinate frames and subscripts to identify specific entities and their components, if needed.
For example, $v^{\mathcal{W}}_{B, \text{x}}$ denote the x-component of the velocity of the robot body ($B$) in the world frame ($\mathcal{W}$).
Fig. 2 shows representative coordinate frames, such as the robot body ($\mathcal{B}$) and platforms ($\mathcal{P}$, $\mathcal{P}_L$, $\mathcal{P}_R$), along with each entity.

For quadruped robots having 12 degrees of freedom (DoF), $\boldsymbol{q}$, $\boldsymbol{\dot{q}}$, $\boldsymbol{\ddot{q}}$, and $\boldsymbol{\tau_q} \in \mathbb{R}^{12}$ represent joint positions, velocities, accelerations, and torques, respectively.
$\boldsymbol{f}_{c, i} \in \mathbb{R}^{3}$ denotes foot contact forces and $c_i \in \{0, 1\}$ are contact indicators for each leg, where $i$ ranges from 0 to 3.

\section{Dynamic Models of Transporters}
\label{sec:tp_types}
Personal transportation platforms, called transporters, encompass devices such as Segways and hoverboards, featuring diverse kinematic variations and control mechanisms ranging from inclination-based to handle-operated systems~\cite{nguyen2004segway, siddhardha2019quadrotor, zapata2024flyboard, hendohover, rosrobosavvybalance2017, hover1hoverboards}. Some further integrate self-balancing controllers that regulate platform inclinations to assist users in maintaining balance.

\begin{figure}[t!]
    \centering 
    \includegraphics[width=\columnwidth]{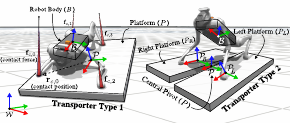}
    \caption{ 
    This figure illustrates the concept of transporter riding tasks involving two types of transporters. Additionally, it introduces some variable notations, such as the coordinate frames for the robot body ($\mathcal{B}$), platform ($\mathcal{P}$), and world ($\mathcal{W}$); entities for the robot body ($B$) and several platforms ($P$, $P_R$, $P_L$); and the foot contact forces ($\mathbf{f}_c$) along with their relative positions ($\mathbf{r}_c$).
    }
    \vspace{-0.5cm}
    \label{figure:platform_types}
\end{figure}

This study investigates two representative transporter types shown in Fig.~\ref{figure:platform_types}.
We focus on transporter dynamics controlled by platform tilts, induced by the robot's weight shifts, given the quadruped robots' limited dexterity, which can only push with their feet. 
We abstract propulsion mechanisms, such as wheels and turbines, into an acceleration-based model along with generalized resistances that emulate ground friction and air resistance.
The specific dynamic models are as follows:

\subsection{Transporter Type 1: Single-Board Design}
Single-board transporters govern forward $\dot{v}_f$ and yaw $\alpha_{P, \text{z}}^{\mathcal{W}}$ accelerations via pitch $\theta_{P, \text{y}}^{\mathcal{W}}$ and roll $\theta_{P, \text{x}}^{\mathcal{W}}$ angles, respectively:
\begin{gather}
    \dot{v}_f = (\dot{v}^{\text{TP}}_{\text{max}}  \textit{clip}(\theta_{P, \text{y}}^{\mathcal{W}}/{\theta^{\text{TP}}_{\text{np}}}) - R(v_f)) / m_P, \\
    v^{\mathcal{W}}_{P, \text{x}} = v_f \cos(\theta^{\mathcal{W}}_{P, \text{z}}), \; v^{\mathcal{W}}_{P, \text{y}} = v_f \sin(\theta^{\mathcal{W}}_{P, \text{z}}), \\
    \!\alpha_{P, \text{z}}^{\mathcal{W}} \!= \!(\alpha^{\text{TP}}_{\text{max}} \textit{sgn}(\theta_{P, \text{y}}^{\mathcal{W}}) \textit{clip}(-\theta_{P, \text{x}}^{\mathcal{W}}/\theta^{\text{TP}}_{\text{np}} ) \!- \!R(\omega_{P, \text{z}}^{\mathcal{W}})) / I_{P, \text{zz}}, \!
\end{gather}
where $\textit{clip}(\cdot)$ returns values clipped to the interval $[-1.0, 1.0]$ and $\textit{sgn}(\cdot)$ outputs $-1.0$ for negative inputs, $1.0$ otherwise. 
$m_P$ is the platform mass, and $I_P$ is its moments of inertia, assuming uniform mass distribution.
$\dot{v}^{\text{TP}}_{\text{max}}$ and $\alpha^{\text{TP}}_{\text{max}}$ represent transporter's maximum forward and angular accelerations, respectively, with $\theta^{\text{TP}}_{\text{np}}$ serving as a normalization parameter. 
$R(v_f)$ and $R(\omega)$ denote generalized resistance forces acting against forward and angular velocities, respectively.
The roll and pitch dynamics, governed by self-balancing controllers and external robot-induced contact forces, are detailed below:
\begin{gather}
    \boldsymbol{\tau}^{\mathcal{P}}_{P} = \textstyle \sum_{i=0}^{3} \mathbf{f}^{\mathcal{P}}_{c, i} \times \mathbf{r}^{\mathcal{P}}_{c, i}, \\
    \alpha_{P, \text{x}}^{\mathcal{P}} = (-k_{p, \text{1}}^{\text{SB}}     \theta_{P, \text{x}}^{\mathcal{W}} - k_{d, \text{1}}^{\text{SB}}      \omega_{P, \text{x}}^{\mathcal{P}} + \tau^{\mathcal{P}}_{P, \text{x}}) / I_{P, \text{xx}}, \\
    \alpha_{P, \text{y}}^{\mathcal{P}} = (-k_{p, \text{2}}^{\text{SB}}      \theta_{P, \text{y}}^{\mathcal{W}} - k_{d, \text{2}}^{\text{SB}}      \omega_{P, \text{y}}^{\mathcal{P}} + \tau^{\mathcal{P}}_{P, \text{y}}) / I_{P, \text{yy}}, 
\end{gather}
where $\bm{k_p}^{\text{SB}}$ and $\bm{k_d}^{\text{SB}} \in \mathbb{R}^2$ denote internal Self-Balancing (SB) gains, and $\mathbf{r}^{\mathcal{P}}_{c, i}$ are foot contact positions on the platform ($\mathcal{P}$).
Please note that the transporter's reactiveness to foot contacts may vary with the transporter's internal parameters and mass.

\begin{figure*}[t]
    \centering 
    \includegraphics[width=1.99\columnwidth]{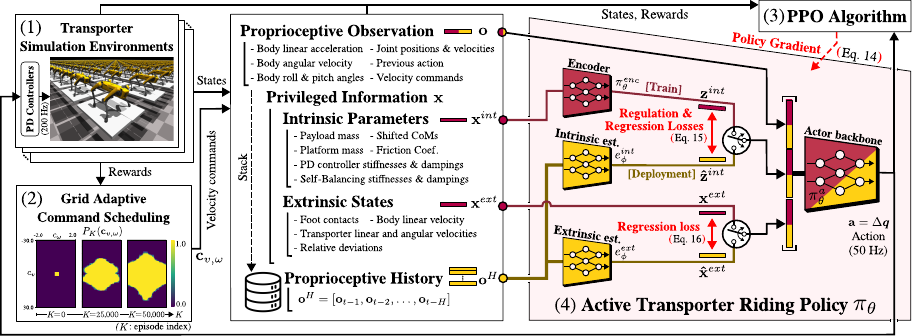}
    \caption{
            \textbf{Overall Framework of the Reinforcement Learning-based Active Transporter Riding Method (\textit{RL-ATR}).} 
            This integrates four key modules for developing a transporter riding policy $\pi_\theta$: 
            (1) simulation environments modeling transporter and robot dynamics; 
            (2) a command scheduling method that systematically raises the riding-task difficulty for effective policy learning; 
            (3) a policy optimization algorithm; and 
            (4) an active transporter riding policy with estimators.
            Components used in the training phase are highlighted in red, those in the deployment phase in yellow, and in both phases in both colors.
    }
    \vspace{-0.6cm}
    \label{figure:overall}
\end{figure*}

\subsection{Transporter Type 2: Two-Board Design}
Two-board transporters consist of two parallel platforms connected by a central pivot ($P$), similar to a bisected single-board design.
Each platform retains one rotational DoF along the y-axis. Therefore, this type-2 design modulates forward and angular accelerations via the average $\theta_{\text{avg}}$ and differential $\theta_{\text{dif}}$ pitch angles of the left and right platforms, respectively:
\begin{gather}
    \theta_{\text{avg}} = (\theta^{\mathcal{W}}_{P_R, \text{y}}   + \theta^{\mathcal{W}}_{P_L, \text{y}})/2, \; \theta_{\text{dif}} = (\theta^{\mathcal{W}}_{P_R, \text{y}}   - \theta^{\mathcal{W}}_{P_L, \text{y}})/2, \\
    \dot{v}_f = (\dot{v}^{\text{TP}}_{\text{max}}       \textit{clip}(\theta_{\text{avg}}/\theta^{\text{TP}}_{\text{np}}  ) - R(v_f))/ (m_{P_L}   + m_{P_R}),\\
    v^{\mathcal{W}}_{P, \text{x}} = v_f     \cos(\theta^{\mathcal{W}}_{P, \text{z}}), \; v^{\mathcal{W}}_{P, \text{y}} = v_f     \sin(\theta^{\mathcal{W}}_{P, \text{z}}),\\
    \!\alpha_{P, \text{z}}^{\mathcal{W}}\! = \!( \alpha^{\text{TP}}_{\text{max}} \textit{sgn}(\theta_{\text{avg}})  \textit{clip}(-\theta_{\text{dif}}/\theta^{\text{TP}}_{\text{np}}  ) \! - \!R(\omega_{P, \text{z}}^{\mathcal{W}}))/I^*_{P, \text{zz}},\!
\end{gather}
where $I^*_{P}$ is the combined inertia of two parallel platforms at the pivot frame $\mathcal{P}$, using the parallel axis theorem~\cite{abdulghany2017generalization}.

Similarly, pitch dynamics are modeled with left and right leg pairs independently controlling their respective platforms:
\begin{gather}
    \boldsymbol{\tau}^{\mathcal{P}_R}_{P_R} = \textstyle \sum_{i=0}^{1} \mathbf{f}^{\mathcal{P}_R}_{c, i} \times \mathbf{r}^{\mathcal{P}_R}_{c, i}, \; \boldsymbol{\tau}^{\mathcal{P}_L}_{P_L} = \textstyle \sum_{i=2}^{3} \mathbf{f}^{\mathcal{P}_L}_{c, i} \times \mathbf{r}^{\mathcal{P}_L}_{c, i},   \\
    \alpha_{P_R, \text{y}}^{\mathcal{P}_R} = (-k_{p, \text{1}}^{\text{SB}}      \theta_{P_R, \text{y}}^{\mathcal{W}} - k_{d, \text{1}}^{\text{SB}}       \omega_{P_R, \text{y}}^{\mathcal{P}_R} + \tau^{\mathcal{P}_R}_{P_R, \text{y}})/I_{P_R, \text{yy}},  \\
    \alpha_{P_L, \text{y}}^{\mathcal{P}_L} = (-k_{p, \text{2}}^{\text{SB}}      \theta_{P_L, \text{y}}^{\mathcal{W}}- k_{d, \text{2}}^{\text{SB}}      \omega_{P_L, \text{y}}^{\mathcal{P}_L} +\tau^{\mathcal{P}_L}_{P_L, \text{y}})/I_{P_L, \text{yy}}.  
\end{gather}

Moreover, we integrate an altitude-maintenance controller, akin to hovering systems~\cite{bouabdallah2007full}, to compensate for the limited controllable DoFs of the two transporter types above. Each provides only two controllable DoFs for forward and turning motions, necessitating a separate altitude control mechanism.

\newpage
    \section{Reinforcement Learning-based \\ Active Transporter Riding method (\textit{RL-ATR})}
We introduce the \textit{RL-ATR} framework (Fig.~\ref{figure:overall}), a RL-based control approach that enables quadruped robots to efficiently navigate long distances utilizing transporters.
The subsequent sections provide a detailed exposition of the \textit{RL-ATR}, covering problem formulation of RL, policy components, reward compositions, a curriculum strategy, and training details.

\subsection{Problem Formulation of RL}
RL aims to develop a policy that maneuvers the transporter to adhere to velocity commands while ensuring the stability of the quadruped robot, accounting for inertia and fictitious inertial forces acting on the robot.
We treat the transporter as part of the environment, which precludes direct control and access to its internal parameters.
Considering the limited data available from the robot's onboard sensors, we formulate this riding problem as a Partially Observable Markov Decision Process (POMDP).
The POMDP is defined by a septuple $( \mathcal{S}, \mathcal{O}, \mathcal{A}, \mathcal{T}, \mathcal{R}, \rho_0, \gamma )$, 
where $\mathcal{S}$ is the state space, $\mathcal{O} \subset \mathcal{S}$ is the observation space, $\mathcal{A}$ is the action space, $\mathcal{T}: \mathcal{S} \times \mathcal{A} \rightarrow \mathcal{S}$ is the state transition function, $\mathcal{R}: \mathcal{S} \times \mathcal{A} \rightarrow \mathbb{R}$ is the reward function, $\rho_0$ is the initial state distribution, and $\gamma \in [0,1)$ is the discount factor. 
At the start of each episode, we initialize\\ the robot with a nominal standing posture $\boldsymbol{q}_0$ at the center of the transporter, represented by $\mathbf{s_0} \sim \rho_0$, with slight randomization of height and joint angles to introduce variability.

We then derive an active transporter riding policy, $\pi_\theta$, by maximizing the expected sum of discounted rewards $J(\pi_\theta)$:
\begin{equation}
\mathbb{E}_{\mathbf{c}_{v, \omega} \sim P(\mathbf{c}_{v, \omega})}
\left[
\mathbb{E}_{\substack{(\mathbf{s}, \mathbf{a}) \sim \rho_{\pi_{\theta}}\\\mathbf{s}_0 \sim \rho_0}}
\left[
\sum_{t=0}^{\infty} \gamma^t  \mathcal{R}(\mathbf{s}_t, \mathbf{a}_t | \mathbf{c}_{v, \omega})
\right]
\right], 
\label{eq:rl_obj}
\end{equation}
where $\theta$ denotes the policy parameters to be optimized, and $\rho_{\pi_\theta}$ is the state-action visitation probability under the policy $\pi_\theta$. 
Here, $\mathbf{c}_{v, \omega}$ represents a set of linear and angular velocity commands sampled from the command distribution $P(\mathbf{c}_{v, \omega})$.\\
Scheduling this command distribution is essential for comprehensive coverage of command spaces (refer to Sec.~\ref{method:CurriculumStrategy}).

Partial observability in POMDPs complicates motor skill acquisition using RL~\cite{zhuang2023robot, gangwani2020learning, meng2021memory}. 
Privileged information $\mathcal{X} \subset \mathcal{S} \setminus \mathcal{O}$, comprising unobservable states, offers valuable environmental context. 
To harness such information, recent works integrate system identification with privileged learning~\cite{Yu2017si, lee2020learning, kumar2021rma, miki2022learning, ji2022concurrent, cheng2024extreme}, transforming POMDPs into MDPs by using simulation-derived data to train policies.
During deployment, estimators substitute the privileged data with estimates derived from a history of observations.
This study employs a regularized online adaptation (ROA) method~\cite{fu2023deep, cheng2024extreme} to enhance policy adaptability against domain variations that affect quadruped robot and transporter dynamics and resolve the situational ambiguity of onboard sensor data captured in the non-inertial frame by inferring robot and transporter velocities with relative deviations, improving transporter-riding performance.

\begin{table}
\vspace{0.2cm}
\centering
\begin{tabular}{l|l|l|l}
\hline
\textbf{NN.} & \textbf{Inputs (dimension)} & \textbf{Hidden Layers} & \textbf{Outputs} \\
\hline 
$\pi_\theta^{a}$ \rule{0pt}{2.2ex} & $\mathbf{o}_t \mid$  $\mathbf{z}^{int}_t \mid$  $\mathbf{x}^{ext}_t$ (75) & [512, 256, 128] & $\mathbf{a}_t$ (12) \\ [0.05cm]
$\pi_\theta^{enc}$ & $\mathbf{x}^{int}_t$ (34) & [128, 64] & $\mathbf{z}^{int}_t$ (16) \\ [0.05cm]
$e^{int}_{\phi}$ & $\mathbf{o}^H$ (\textit{H} x 46) & $\text{CNN-GRU}$ + [128] & $\mathbf{\hat{z}}^{int}_t$ (16) \\ [0.05cm]
$e^{ext}_{\phi}$ & $\mathbf{o}^H$ (\textit{H} x 46) & $\text{CNN-GRU}$ + [128] & $\mathbf{\hat{x}}^{ext}_t$ (13) \\
\hline
\end{tabular}
\caption{
Neural Network (NN.) architectures of the \textit{RL-ATR} framework: the actor backbone $\pi_\theta^{a}$, encoder $\pi_\theta^{enc}$, and both intrinsic $e_\phi^{int}$ and extrinsic $e_\phi^{ext}$ estimators. 
Vertical bars ($\mid$) signify the concatenation of input features, and square brackets ([$\cdot$]) represent Multi-Layer Perceptron (MLP) layers.
The CNN-GRU, combining a Convolutional Neural Network with a Gated Recurrent Unit, processes time-dependent features. $H$ is the history length.
}
\label{tab:network_architecture}
\vspace{-0.6cm}
\end{table}

\subsection{Active Transporter Riding Policy} 
Following the problem formulation of RL, we detail policy and reward compositions within the \textit{RL-ATR} framework. 
As illustrated in Fig.~\ref{figure:overall}, an active transporter riding policy $\pi_\theta$ comprises an actor backbone $\pi_\theta^{a}$ and an encoder $\pi_\theta^{enc}$. 
It also integrates intrinsic and extrinsic estimators, $e_\phi^{int}$ and $e_\phi^{ext}$, for system identification, where $\phi$ denotes estimator parameters. 
The network architectures are further detailed in TABLE~\ref{tab:network_architecture}.

\subsubsection{\textbf{Policy Output}} At each time step, the policy generates joint displacements $\Delta \boldsymbol{q}$, deviating from the nominal standing posture $\boldsymbol{q}_0$, as actions $\mathbf{a} \in \mathcal{A}$. Proportional-Derivative (PD) controllers then generate torques $\boldsymbol{\tau_q}$ using $\Delta \boldsymbol{q} + \boldsymbol{q}_0$ as targets.

\begin{table}[t]
\centering
\caption{Domain Randomization Intrinsic Parameters, $\mathbf{x}^{int} \in \mathbb{R}^{34}$}
\renewcommand{\arraystretch}{1.2}
\begin{tabular}{c|c|c|c}
\hline
\textbf{Term} & \textbf{Training Range} & \textbf{Testing Range} & \textbf{Unit}\\
\hline
\multicolumn{4}{c}{Quadruped Robots (PD: joints' PD controllers)} \\
\hline
Payload Mass & $[\;0.0, 1.0\;]$         & $[\;0.0, 3.0\;]$ & \SI{}{\kilogram}\\
Shifted CoM  & $[\,-0.2, 0.2\;]^{3}$    & $[\,-0.25, 0.25\;]^{3}$ & \SI{}{\meter}\\
PD Stiffness             & $[\;36, 44\;]^{12}$      & $[\;32, 48\;]^{12}$ & -\\
PD Damping                & $[\;0.8, 1.2\;]^{12}$    & $[\;0.6, 1.4\;]^{12}$ & -\\
\hline
\multicolumn{4}{c}{Transporters (SB: internal Self-Balancing controller)} \\
\hline
Platform Mass                 & $[\;-0.5, 0.5\;]$         & $[\;-1.0, 1.0\;]$ & \SI{}{\kilogram}\\
Friction Coef.                & $[\;0.8, 1.2\;]$         & $[\;0.7, 1.5\;]$ & -\\
SB Stiffness ($\bm{k_p}^{\text{SB}}$) & $[\;0.8, 1.5\;]^2$      & $[\;0.5, 2.0\;]^2$ & -\\ 
SB Damping ($\bm{k_d}^{\text{SB}}$)   & $[\;0.02, 0.03\;]^2$      & $[\;0.01, 0.05\;]^2$ & -\\ [0.02cm]
\hline
\end{tabular}
\vspace{-0.55cm}
\label{table:domainparams}
\end{table}

\subsubsection{\textbf{Policy Input}} The policy $\pi_\theta$ makes use of distinct input sources during training and deployment phases, as marked by red and yellow colors in Fig.~\ref{figure:overall}, respectively. 
In developing riding skills, the policy takes a proprioceptive observation $\mathbf{o} \in \mathcal{O}$ and the privileged information $\mathbf{x} \in \mathcal{X}$ as inputs.

The proprioceptive observation $\mathbf{o}$ is composed of sensor measurements $\mathbf{o}_m$, the previous action $\mathbf{a}_{t-1}$, and the velocity command $\mathbf{c}_{v, \omega}$, such that $\mathbf{o} = [\mathbf{o}_m, \mathbf{a}_{t-1}, \mathbf{c}_{v, \omega}]$.
Here, $\mathbf{o}_m = [\boldsymbol{\dot{v}}^{\mathcal{B}}_{B}, \boldsymbol{\omega}^{\mathcal{B}}_{B}, \boldsymbol{\theta}^{\mathcal{W}}_{B, \text{xy}}, \boldsymbol{q}, \boldsymbol{\dot{q}}]$ includes the body's linear acceleration, angular velocity, orientations along with joint positions and velocities. 
For brevity, we omit the current time notation $t$.

We bifurcate the privileged information into intrinsic and extrinsic components $\mathbf{x} = [\mathbf{x}^{int}, \mathbf{x}^{ext}]$. 
The intrinsic component $\mathbf{x}^{int} \in \mathcal{X}^{int}$ captures dynamic model parameters, as listed in TABLE~\ref{table:domainparams}.
These properties cause varying environmental responses to identical actions, potentially hindering performance if not considered~\cite{margolis2023walk, margolis2024rapid}.
We incorporate this intrinsic information via an intrinsic latent vector $\mathbf{z}^{int} \in \mathbb{R}^{16}$, embedded using the encoder $\pi_\theta^{enc}$.
The extrinsic component $\mathbf{x}^{ext} = [(c_0, c_1, c_2, c_3), \boldsymbol{v}^{\mathcal{B}}_{B}, \boldsymbol{v}^{\mathcal{B}}_{P}, \boldsymbol{\omega}^{\mathcal{B}}_{P}, \boldsymbol{p}^{\mathcal{P}}_{B, \text{xy}}, \theta^{\mathcal{P}}_{B, \text{z}}] \in \mathcal{X}^{ext}$ includes robot and transporter states, comprising foot-contact indicators; body and platform velocities in the body frame $\mathcal{B}$; and the robot's relative pose on the platform.
This information enhances the policy's ability to maneuver and maintain balance by recognizing the spatial relationship between the robot and platform and interpreting reference frame motions. This awareness is essential for maintaining or regaining the equilibrium of the robot in the non-inertial transporter frame.

\subsubsection{\textbf{Estimators}}
To bridge the information gap between the training and deployment, we concurrently develop the intrinsic and extrinsic estimators, $e_\phi^{int}$ and $e_\phi^{ext}$, with the policy.
These estimators infer the leveraged privileged information, $\mathbf{x}^{int}$ and $\mathbf{x}^{ext}$, using historical proprioceptive observations $\mathbf{o}^H = [\mathbf{o}_{t-1}, \mathbf{o}_{t-2}, \ldots, \mathbf{o}_{t-H}] \in \mathcal{O}^H$. 
Each estimator maps the historical observations $\mathbf{o}^H$ to its respective targets: 
the intrinsic estimator $e_\phi^{int}$ infers the latent vector $\mathbf{\hat{z}}^{int} \in \mathbb{R}^{16}$ that represents the embedded intrinsic properties $\mathbf{z}^{int}$. 
The extrinsic estimator $e_\phi^{ext}$ explicitly deduces the extrinsic component $\mathbf{\hat{x}}^{ext} \in \mathcal{X}^{ext}$ to approximate the true extrinsic states $\mathbf{x}^{ext}$.
As noted in~\cite{kumar2021rma}, transferring privileged information in the latent space improves adaptation performance. 
In contrast, explicit inference provides explainability and facilitates sensor fusion, potentially improving measurement accuracy.

Both estimators are simultaneously trained with the policy, optimized with Eq.~\ref{eq:rl_obj}, using the following regression losses:

\begin{align}
L^{int} &= \| \mathbf{\hat{z}}^{int} - sg[\mathbf{z}^{int}] \|_2^2 + \lambda \| sg[\mathbf{\hat{z}}^{int}] - \mathbf{z}^{int} \|_2^2, \\
L^{ext} &= \| \mathbf{\hat{x}}^{ext} - \mathbf{x}^{ext} \|_2^2, 
\label{eq:estloss}
\end{align}
where $sg[\cdot]$ is the stop-gradient operator and $\lambda$ is a regularization weight that helps mitigate the reality gap~\cite{fu2023deep, cheng2024extreme}.

\begin{table}[t!]
\centering
\begin{threeparttable}
\caption{Reward Composition for Transporter (TP) Riding Skills \\
(For a more detailed description, please refer to Sec.~\ref{label:RewardFunction}.)}
\label{table:rewardfunctions}
\renewcommand{\arraystretch}{1.2}
\begin{tabular}{r|l}
\hline
\multicolumn{1}{c|}{\textbf{Reward Term}} & \multicolumn{1}{c}{\textbf{Expression}} \\
\hline
\multicolumn{2}{c}{Task Rewards: $\mathcal{R}^{\text{task}} = \sum_{i=0}^{8} r_i$} \\
\hline
Forward Command ($r_0$) & $k_0 \exp(-\| p^{\mathcal{P}_{\text{planar}}}_{P, \text{x}} - \text{c}_{v} \|_2 / 0.5)$ \\
Steering Command ($r_1$) & $k_1 \exp(-\| \omega^{\mathcal{P}_{\text{planar}}}_{P, \text{z}} - \text{c}_{\omega} \|_2 / 0.5)$ \\
Position Alignment ($r_2$) & $k_2 \| \boldsymbol{p}^{\mathcal{W}}_{B, \text{xy}} - \boldsymbol{p}^{\mathcal{W}}_{P, \text{xy}} \|_2$ \\
Heading Alignment ($r_3$) & $k_3 \| \theta^{\mathcal{W}}_{B, \text{z}} - \theta^{\mathcal{W}}_{P, \text{z}} \|_2$ \\
CoM Stabilization ($r_4$) & $-k_4 \mathds{1}_{\text{outside-foot-polygon}}(\boldsymbol{p}^{\mathcal{W}}_{\textit{CoM}, \text{xy}})$ \\
ZMP Stabilization ($r_5$) & $-k_5 \mathds{1}_{\text{outside-foot-polygon}}(\boldsymbol{p}^{\mathcal{W}}_{\textit{ZMP}, \text{xy}})$ \\
Contact Maintenance ($r_6$) & $-k_6 (4 - \sum_{i=0}^{3} c_i)$ \\
Height Maintenance ($r_7$) & $-k_7 \| (p^{\mathcal{W}}_{B, \text{z}} - p^{\mathcal{W}}_{P, \text{z}}) - h_{\text{des}} \|_2$ \\
TP Smoothness ($r_8$) & $-k_8 ( \|\boldsymbol{\dot{v}}^{\mathcal{P}}_{P}\|_2 + \|\boldsymbol{\alpha}^{\mathcal{P}}_{P}\|_2 )$ \\
\hline
\multicolumn{2}{c}{Regularization Rewards: $\mathcal{R}^{\text{reg}} = \sum_{i=9}^{17} r_i$} \\
\hline
Body Orientation ($r_{9}$) & $- k_9 \| \boldsymbol{\theta}^{\mathcal{W}}_{B, \text{xy}} \|_2$ \\
Body Velocity ($r_{10}$) & $- k_{10} ( \| \boldsymbol{\omega}^{\mathcal{B}}_{B, \text{xy}} \|_2 + | v^{\mathcal{W}}_{B, \text{z}} | )$ \\
Action Smoothness ($r_{11}$) & $- k_{11} \| \mathbf{a} - \mathbf{a}_{t-1} \|_2$ \\
Joint Smoothness ($r_{12}$) & $- k_{12} \| \boldsymbol{\tau_q} \|_2 - k_{13} \| \boldsymbol{\dot{q}} \|_2 - k_{14} \| \boldsymbol{\ddot{q}} \|_2$ \\
Postural Deviation ($r_{13}$) & $- k_{15} (\| \boldsymbol{q} - \boldsymbol{q}_0 \|_2)$ \\
Energy Efficiency ($r_{14}$) & $- k_{16} \sum_{j=0}^{11} \max( \tau_{q}[j] \dot{q}[j], 0.0)$ \\
Force Regulation ($r_{15}$) & $- k_{17} \sum_{i=0}^{3} \max(\|\bm{f}_{c, i}\|_2 - f_{\text{tol}}, 0.0)$ \\
Collision Avoidance ($r_{16}$) & $- k_{18} \mathds{1}_{\text{collision}}$ \\
Termination ($r_{17}$) & $- k_{19} \mathds{1}_{\text{termination}}$ \\
\hline
\end{tabular}
\begin{tablenotes}[flushleft]
    \scriptsize 
    \item \textbullet\; $\mathcal{P}_{\text{planar}}$: The platform frame with zero roll and pitch angles.
    \item \textbullet\; $h_{\text{des}}$: Desired body height \quad \textbullet\; $f_{\text{tol}}$: Tolerated maximum contact force.
    \item \textbullet\; $k_{0}, \dots, k_{19}$: Non-negative reward function weights.
\end{tablenotes}
\end{threeparttable}
\vspace{-1.15cm}
\end{table}

\subsubsection{\textbf{Reward Composition}}
\label{label:RewardFunction}
We design the reward function $\mathcal{R}$ to enable the policy $\pi_\theta$ to safely maneuver transporters in response to the velocity command $\mathbf{c}_{v, \omega} = [c_v, c_\omega] \in \mathbb{R}^{2}$. 
The total reward is the summation of task and regularization rewards, $\mathcal{R} = \mathcal{R}^{\text{task}} + \mathcal{R}^{\text{reg}}$, as enumerated in TABLE~\ref{table:rewardfunctions}.

The task rewards $\mathcal{R}^{\text{task}} = \sum_{i=0}^{8} r_i$ address key aspects of the riding task: 
$r_0$ and $r_1$ ensure the transporter adheres to commanded velocities; 
$r_2$ and $r_3$ align the center positions and orientations between the robot and transporter; 
$r_4$ and $r_5$ guide to ensure static stability by keeping the CoM and Zero Moment Point (ZMP) within the polygon defined by the foot positions; 
$r_6$ encourages foot contacts to effectively transmit contact forces and generate frictional forces that counteract inertial forces; 
$r_7$ prevents the robot from lying down on the transporter; and 
$r_8$ slightly mitigates stability issues due to inertia effects by penalizing abrupt transporter accelerations.

Training a policy solely on task rewards can lead to local minima and unexpected motions \cite{margolis2023walk}. 
To mitigate this issue, we integrate regularization rewards $\mathcal{R}^{\text{reg}} = \sum_{i=9}^{17} r_i$: 
$r_9$ and $r_{10}$ regulate body tilts and velocities; 
$r_{11}$ and $r_{12}$ promote smooth joint movements; 
$r_{13}$ minimizes deviations from the nominal posture; 
$r_{14}$ reduces joint motor power usage; 
$r_{15}$ penalizes excessive contact forces to protect hardware; 
$r_{16}$ and $r_{17}$ prevent the policy from entering unsafe states.
We terminate episodes early if the robot risks flipping or falling off the transporter.
This strategy enhances learning efficiency by reducing wasteful exploration of unfeasible states~\cite{peng2018deepmimic}.

 \begin{figure*}[ht!]
    \centering 
    \includegraphics[width=2\columnwidth]{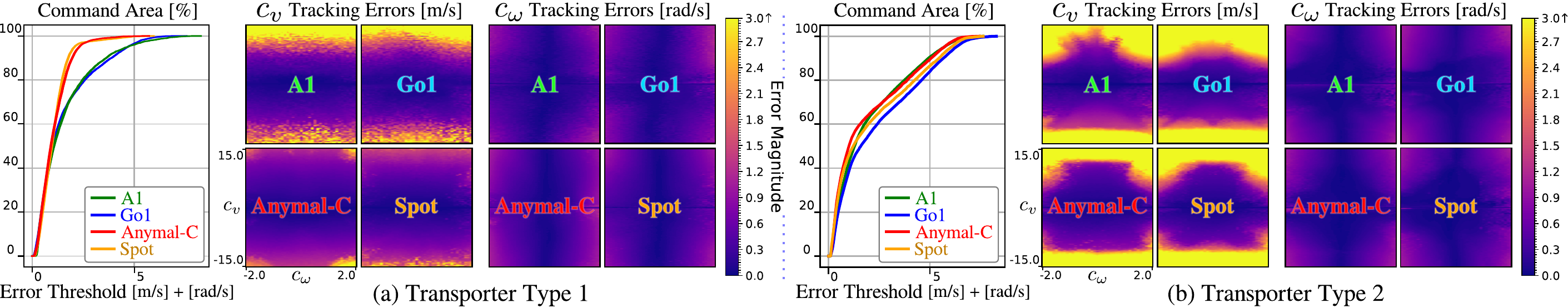}
    \vspace{-0.1cm} 
    \caption{ 
    Heatmaps of tracking errors for $c_v$ (forward velocity) and $c_\omega$ (yaw rate) commands on the $\mathcal{C}^{\text{eval}}_{v, \omega}$, with corresponding command area graphs~\cite{margolis2024rapid}.  
    }
    \label{fig:fig5}
    \vspace{-0.55cm}
\end{figure*} 
\subsection{Curriculum Strategy}
\label{method:CurriculumStrategy}
Learning complex motor skills from scratch is challenging, particularly in transporter riding tasks. Initial random policies often fail to track high-velocity commands due to intricate transporter dynamics and balancing demands, such as standing on inclined platforms and managing fictitious inertial forces.  
Moreover, the greater the robot's momentum, the greater the external force required for velocity adjustments.
Consequently, these multifaceted challenges make meaningful rewards hard to obtain, hindering the learning process.

Therefore, we implement a grid adaptive update rule~\cite{margolis2024rapid}, progressively expanding the command distribution $P(\mathbf{c}_{v, \omega})$ according to the maturity of the riding ability.
The rule raises the probability of adjacent regions of the sampled command, $\mathbf{c}_{v, \omega}^{\Delta} \in \mathbf{c}_{v, \omega} \oplus \Delta$, when tracking rewards surpass thresholds:
\vspace{-0.15cm}
\begin{equation}
P_{K+1}(\mathbf{c}_{v, \omega}^{\Delta}) =
\begin{cases}
\; P_K(\mathbf{c}_{v, \omega}^{\Delta}), \hspace{1.342cm} \text{if } r_0 < \gamma_v \vee r_1 < \gamma_\omega, \\
\; \max(P_K(\mathbf{c}_{v, \omega}^{\Delta}) + \delta, \; 1.0), \hspace{0.8cm} \text{otherwise},
\end{cases}
\end{equation}
where $\oplus$ is the Minkowski sum operator, $r_0$, $r_1$ are tracking rewards for the command $\mathbf{c}_{v, \omega}$ as defined in TABLE~\ref{table:rewardfunctions}, $\gamma_v$, $\gamma_\omega$ are the corresponding thresholds, $K$ is the episode index, $\Delta$ is expansion regions, and $\delta$ is the probability increment.
The distribution is initialized with a small range of velocities:
\vspace{-0.15cm}
\begin{equation}
P_0(\mathbf{c}_{v, \omega}) =
\begin{cases}
\frac{1}{4 c_v^{\text{init}} c_\omega^{\text{init}}}, & \text{if } \mathbf{c}_{v, \omega} \in [-c_v^{\text{init}}, c_v^{\text{init}}] \times [-c_\omega^{\text{init}}, c_\omega^{\text{init}}], \\
0, & \text{otherwise},
\end{cases}
\end{equation}
where $c_v^{\text{init}}$, $c_\omega^{\text{init}}$ define initial command ranges.
Fig.~\ref{figure:overall} exhibits how the distribution $P_K(\mathbf{c}_{v, \omega})$ expands over episodes $K$.

\subsection{Training Details}
\label{sec:3-C-6}
We utilized Isaac Gym~\cite{makoviychuk2021isaac} to operate 4,096 environments concurrently, each featuring a robot and a transporter with randomly sampled intrinsic properties.
To enhance policy robustness against external perturbations and sudden command changes, we applied random forces to the robot and platforms at \SI{3}{\s} intervals and resampled the commands $\mathbf{c}_{v, \omega}$ every \SI{5}{\s}.

We optimized the riding policy $\pi_\theta$ adopting the Proximal Policy Optimization (PPO)~\cite{schulman2017proximal} as per the RL objective function in Eq.~\ref{eq:rl_obj}, while also minimizing system identification losses in Eqs.~15 and~16.
We designed the policy $\pi_\theta$ to be stochastic for state exploration, drawing outputs from a diagonal Gaussian distribution with means derived from the actor backbone $\pi_{\theta}^a$ and standard deviations parameterized by $\boldsymbol{\theta}^\text{std} \in \mathbb{R}^{12}$.
As for the hyperparameters, we empirically determined the effective values:
$H = 10$ (corresponding to a \SI{0.2}{\s} history); $k_{0,1,\ldots,19} = $ [$8.0$, $8.0$, $30.0$, $4.0$, $1.0$, $1.0$, $2.0$, $1.0$, $1.0$, $0.9$, $10^{-3}$, $10^{-5}$, $10^{-4}$, $10^{-4}$, $10^{-7}$, $10^{-2}$, $10^{-4}$, $10^{-2}$, $10.0$, $10.0$]; $h_{\text{des}}$ depends on the robot models; $f_{\text{tol}} = 100$ N; and $\lambda = 0.2$.
The scheduling parameters are $\Delta = \{\mathbf{c}_{v, \omega} \in \mathbb{R}^2 : |c_v| \leq 0.2, |c_\omega| \leq 0.2\}$, representing a square region in the command space; $\delta = 0.1$; $\gamma_v$ and $\gamma_\omega$ set at \SI{80}{\percent} of their maximum values; $c_v^{\text{init}} = 0.5$; and $c_\omega^{\text{init}} = 0.3$.

The policy $\pi_{\theta}$ converged after around 75,000 episodes $K$, with each generating \SI{10.0}{\s} of data from all environments. 
This entire process took about 72 hours on a desktop with an RTX 4090 GPU, an Intel i9-9900K CPU, and 64GB RAM.

\begin{table}[t]
\vspace{+0.22cm}
\centering
\renewcommand{\arraystretch}{1.2}
\begin{tabular}{c|c|c|c}
\hline
\textbf{Group} & \textbf{Robot Model} & \textbf{Dimension (m)} & \textbf{Mass (kg)} \\
\hline
\multirow{2}{*}{G1} & A1       & $[0.50 \times 0.30 \times 0.40]$ & $11.74$ \\
 & Go1 & $[0.65 \times 0.28 \times 0.40]$ & $12.14$ \\
 \hline
\multirow{2}{*}{G2} & Anymal-C & $[0.93 \times 0.53 \times 0.89]$ & $43.51$ \\
& Spot & $[1.10 \times 0.50 \times 0.61]$ & $32.60$ \\
\hline
\end{tabular}
\caption{
To evaluate transporter compatibility, we group robots by size: A1 and Go1 are in Group 1, and Anymal-C and Spot are in Group 2. We set transporter dimensions as $[0.9 \times 0.7 \times 0.05]$ (m) for G1 and $[1.5 \times 1.1 \times 0.05]$ (m) for G2, along with masses of \SI{11.5}{\kilogram} and \SI{30}{\kilogram}, respectively.
}
\label{tab:experiment_configs}
\vspace{-0.6cm}
\end{table}

\section{Experimental Results}
\label{sec:4}
To corroborate the effectiveness of the \textit{RL-ATR}, we assess command tracking accuracy and navigation efficiency, along with a detailed verification of each component's contribution.

\begin{table*}[t!]
\centering
\caption{Estimation Accuracy of Intrinsic ($e^{int}_{\phi}\!:\mathbf{o}^H \! \rightarrow  \mathbf{\hat{z}}^{int}$) and Extrinsic ($e^{ext}_{\phi}\!:\mathbf{o}^H \! \rightarrow  \mathbf{\hat{x}}^{ext}$) Estimators 
} 
\vspace{-0.1cm}
\renewcommand{\arraystretch}{1.25}
\resizebox{2\columnwidth}{!}{
    \begin{tabular}{c|cccc|ccc|ccc|ccc|cc|c}
    \hline 
    \multirow{2}{*}{\begin{tabular}{@{}c@{}} Intrinsic \\ [-0.07cm] Latent Vector: \\ [-0.03cm] $|| \mathbf{\hat{z}}^{int} - \mathbf{z}^{int} ||_2$\end{tabular}} & 
    \multicolumn{16}{c}{Extrinsic States: $|| \mathbf{\hat{x}}^{ext}[i] - \mathbf{x}^{ext}[i]||_1 (i = [0, 1, \ldots, 15])$} \\
    \cline{2-17}
     & 
    \multicolumn{4}{c|}{\begin{tabular}{@{}c@{}}Contact States \\ [-0.07cm] $(c_0, c_1, c_2, c_3) \in \mathbb{R}^{4}$\end{tabular}} & 
    \multicolumn{3}{c|}{\begin{tabular}{@{}c@{}}Body Linear Velocity \\ [-0.07cm] $\boldsymbol{v}^{\mathcal{B}}_{B} \in \mathbb{R}^{3}$\end{tabular}} & 
    \multicolumn{3}{c|}{\begin{tabular}{@{}c@{}}Transporter Linear Velocity \\ [-0.07cm] $\boldsymbol{v}^{\mathcal{B}}_{P} \in \mathbb{R}^{3}$\end{tabular}} & 
    \multicolumn{3}{c|}{\begin{tabular}{@{}c@{}}Transporter Angular Velocity \\ [-0.07cm] $\boldsymbol{\omega}^{\mathcal{B}}_{P} \in \mathbb{R}^{3}$\end{tabular}} &
    \multicolumn{2}{c|}{\begin{tabular}{@{}c@{}}Relative Position \\ [-0.07cm] $\boldsymbol{p}^{\mathcal{P}}_{B, \text{xy}} \in \mathbb{R}^{2}$\end{tabular}}&
    \multicolumn{1}{c}{\begin{tabular}{@{}c@{}}Relative Orientation \\ [-0.07cm] $\theta^{\mathcal{P}}_{B, \text{z}} \in \mathbb{R}$\end{tabular}} \\
    \hline
    $0.0195$ & $0.0170$ & $0.0610$ & $0.0246$ & $0.1035$ & $0.1128$ & $0.0557 $& $0.1073 $& $0.1074$ & $0.0564$ & $0.0124 $& $0.0112$ & $0.0141$ & $0.0461$ & $0.0361$ & $0.0283$ & $0.0013$ \\ [-0.07cm]
    $\pm 0.0099$ & $\pm0.0085$ & $\pm0.0293$ & $\pm0.0109$ & $\pm0.1608$ & $\pm0.0976$ & $\pm0.0155$ & $\pm0.1653$ & $\pm0.1016$ & $\pm0.0124$ & $\pm0.0003$ & $\pm0.0003$ & $\pm0.0003$ & $\pm0.0017$ & $\pm0.0017$ & $\pm0.0049$ & $\pm0.0008$ \\
    \hline
    \end{tabular}
    }
\vspace{-0.25cm}
\label{table:table5}
\end{table*}

\subsection{Configuration of Transporters}
We configured transporter dynamics to achieve maximum forward and angular accelerations of \SI{12}{\meter/\second\squared} and \SI{3}{\radian/\second\squared} at \SI{45}{\degree} angles, with $\dot{v}^{\text{TP}}_{\text{max}} = 12$, $\alpha^{\text{TP}}_{\text{max}} = 3$, and $\theta^{\text{TP}}_{\text{np}} = \SI{0.78}{\radian}$.\\
We modeled the resistance as $R(x) = 0.2 + 0.05x + 0.005x^2$ for both forward and angular velocities.
Additionally, we defined transporter specifications to validate cross-robot compatibility of the same transporters, as detailed in TABLE IV.

\subsection{Evaluation of Transporter Riding Ability}
\label{exp:command_tracking}
We examined eight combinations of two transporter types and four robot models (A1, Go1, Anymal-C, and Spot~\cite{taheri2023study}) to comprehensively evaluate the applicability of the \textit{RL-ATR}.
For each combination, we generated 10,000 environments with randomly sampled intrinsic properties within test ranges (TABLE~\ref{table:domainparams}).
We measured command tracking errors over \SI{10}{\s} interval for each grid point on an evaluation command space $\mathcal{C}^{\text{eval}}_{v, \omega} = [-15.0, 15.0] \times [-2.0, 2.0]$ with \SI{0.1}{} resolution.

Fig. 4 presents root-mean-square tracking error heatmaps for the evaluation command space $\mathcal{C}^{\text{eval}}_{v, \omega}$, alongside command area graphs~\cite{margolis2024rapid}. 
The command area denotes the command space portion where the policy tracks commands within an error threshold. 
The \textit{RL-ATR} demonstrates proficient riding skills across various robot-transporter combinations, covering a range of the command space.
We also confirmed transporter compatibility, as robots within the same group adeptly managed the same transporter despite their kinematic differences.

Tracking performance drops in high-velocity regions due to increased inertial and resistance forces. 
Notably, group-1 robots with type-1 transporters demonstrate deteriorated performance under high-velocity commands because they have insufficient mass to generate adequate platform-tilting forces.  
Meanwhile, type-2 transporters exhibit inferior performance compared to type-1, due to intricate maneuvering challenges associated with their dual-platform operational mechanisms.

\begin{figure}[t]
    \centering 
    \vspace{-0.2cm}
    \includegraphics[width=\columnwidth]{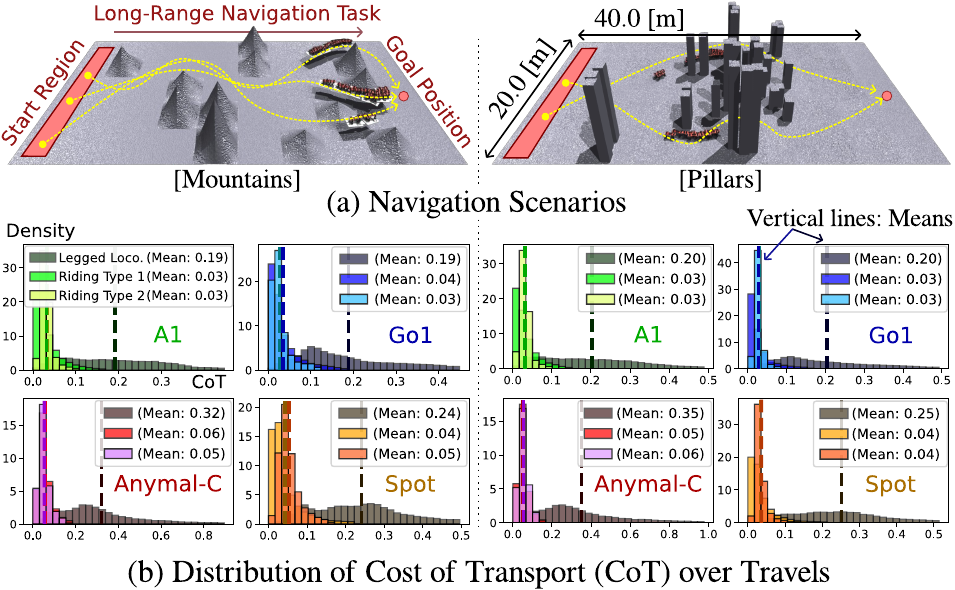}
    \caption{
        \textbf{Long-range Navigation Efficiency Analysis}.
        (a) Two experimental scenarios, with yellow dotted lines illustrating representative planned paths. 
        (b) Distributions of the mechanical Cost of Transport (CoT)~\cite{bjelonic2018skating} for legged locomotion~\cite{margolis2024rapid} and riding approaches using two types of transporters.
    }
    \vspace{-0.5cm}
\end{figure}

\subsection{Evaluation of Long-Range Navigation Efficiency}
\label{sec:navefficiency}
To assess transporter usage efficiency in long-range travel, we set up two environments (Fig. 5-(a)) and generated fifty traversable paths using spline-based RRT~\cite{yang2014spline} for randomly selected start positions.
We then evaluated the mechanical Cost of Transport (CoT)~\cite{bjelonic2018skating} of legged locomotion~\cite{margolis2024rapid} and riding approaches.
To ensure a fair comparison, each method traversed identical paths at consistent speeds (\SI{1.5}{\meter/\second} for G1 and \SI{3}{\meter/\second} for G2) and successfully reached a goal position.
The CoT, a dimensionless power usage metric, is defined as: $\mathbb{E}_{t, j} [ \max(\tau_q[j] \dot{q}[j], 0)/(m  g  v_{\text{avg}})]$, where $m$ is robot mass, $g$ is gravitational acceleration, and $v_{\text{avg}}$ is average travel speed.

Fig. 5 shows CoT distributions over trips driven by a pure pursuit algorithm~\cite{samuel2016review}.
Transporters significantly reduced the robots' power consumption across all robot-transporter pairs by allowing robots to harness the transporter's driving forces, requiring only maneuvering and balancing efforts over travel.

\subsection{Analysis of Components within the \textit{RL-ATR}}
To assess the viability of inferring privileged information from historical observations, we evaluated the intrinsic and extrinsic estimators. TABLE~\ref{table:table5} shows the prediction accuracy of each component, measured during a 10-second command tracking evaluation described in Sec.~\ref{exp:command_tracking}. 
These relatively low prediction errors validate the feasibility of this system identification approach. 
Fig. 6 further displays the prediction results for the continuously changing transporter velocity in response to a manually instructed command sequence.

Furthermore, we examined the contributions of the command curriculum strategy and the utilization of intrinsic and extrinsic transporter information via estimators.
We trained the policies following the same procedure outlined in Sec. IV, excluding ablation components.
Fig. 7 shows command area graphs and combined tracking error heatmaps for each experiment within the evaluation command space $\mathcal{C}^{\text{eval}}_{v, \omega}$. 
The converged policy without the command scheduling scheme failed to track commands, and a lack of transporter information resulted in limited coverage of the command space due to unclear situational awareness in the non-inertial frames.

The attached video intuitively demonstrates the results.

\begin{figure}[t!]
    \centering 
    \includegraphics[width=\columnwidth]{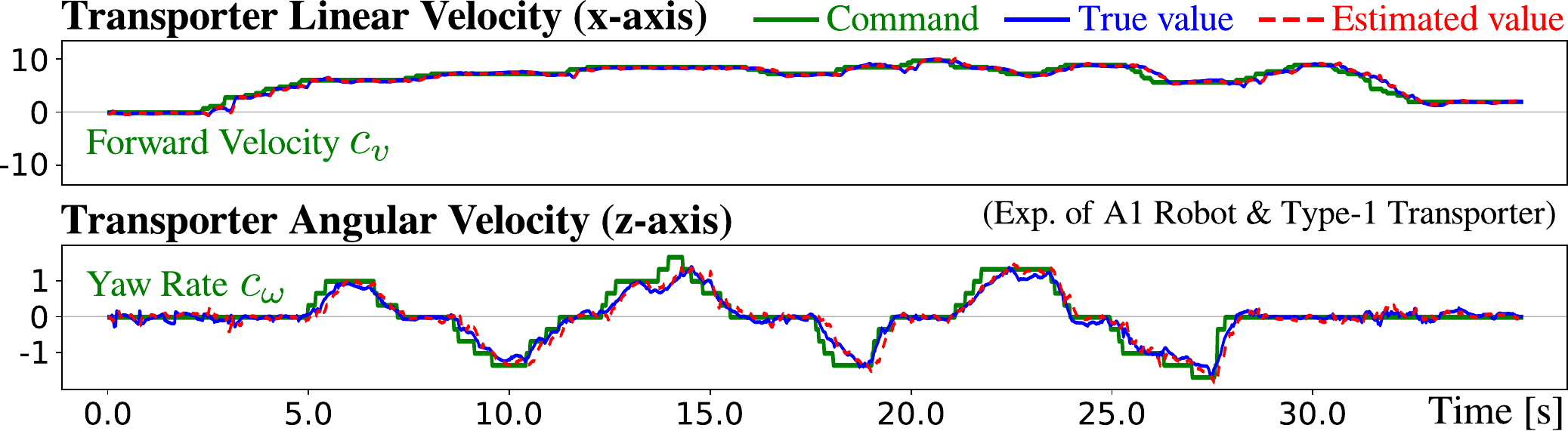}
    \caption{
    Illustrations of command tracking and transporter-velocity estimation accuracy over the course of a manually instructed command sequence.
}
        \label{fig:fig7}
        \vspace{-0.2cm}
\end{figure}

\begin{figure}[t]
    \centering 
    \includegraphics[width=\columnwidth]{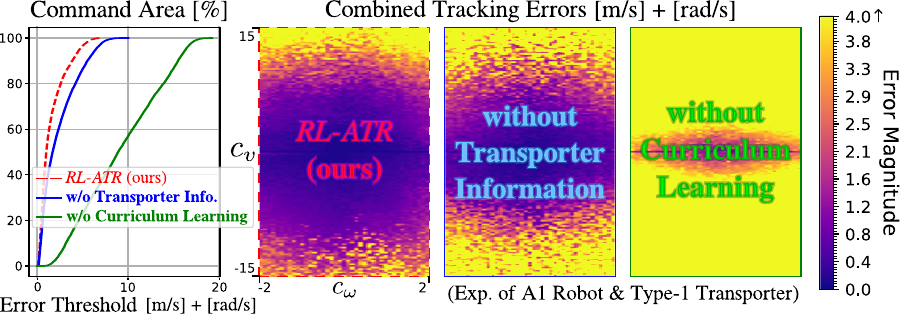}
    \vspace{-0.5cm}
    \caption{ 
        \textbf{Ablation Study.} 
        Heatmaps and command area graphs of combined tracking errors for forward ($c_v$) and angular ($c_\omega$) velocity commands.
        Due to space limits, we include results only for the A1 robot and type-1 transporter. 
    }
        \vspace{-0.5cm}
\end{figure}

    \section{Conclusion}
\label{sec:5}
We introduced \textit{RL-ATR}, a low-level controller enabling quadruped robots to utilize personal transporters for efficient long-range navigation. 
Through comprehensive experiments, we demonstrated the feasibility of RL in developing proficient riding skills for distinct transporter dynamics along with cross-robot compatibility of transporters.
Future work includes real-world validation with physical transporters. We also plan to incorporate mounting and dismounting capabilities for seamless transitions, along with exteroceptive sensors for autonomous navigation in complex environments.

\section*{Acknowledgments}
\footnotesize
This work was supported by the Institute of Information \& Communications Technology Planning \& Evaluation (IITP) and the National Research Foundation of Korea (NRF) grants, funded by the Korea government (MSIT) (No. RS-2023-00237965, RS-2023-00208506).

    {
        \small
        \bibliographystyle{ieee_format/ieee}
        \bibliography{./ref}

@article{arm2023scientific,
  title={Scientific exploration of challenging planetary analog environments with a team of legged robots},
  author={Arm, P. and Waibel, G. and Preisig, J. and others},
  journal={Science Robotics},
  volume={8},
  number={80},
  pages={eade9548},
  year={2023},
  publisher={American Association for the Advancement of Science}
}

@article{lindqvist2022multimodality,
  title={Multimodality robotic systems: Integrated combined legged-aerial mobility for subterranean search-and-rescue},
  author={Lindqvist, B. and others},
  journal={Robotics and Autonomous Systems},
  volume={154},
  pages={104134},
  year={2022},
  publisher={Elsevier}
}

@article{delmerico2019current,
  title={The current state and future outlook of rescue robotics},
  author={Delmerico, J. and others},
  journal={Field Robotics},
  volume={36},
  number={7},
  pages={1171--1191},
  year={2019},
  publisher={Wiley Online Library}
}

@article{bellicoso2018advances,
  title={Advances in real-world applications for legged robots},
  author={Bellicoso, C. D. and others},
  journal={Field Robotics},
  volume={35},
  number={8},
  pages={1311--1326},
  year={2018},
  publisher={Wiley Online Library}
}

@article{lu2023whole,
  title={Whole-body motion planning and control of a quadruped robot for challenging terrain},
  author={Lu, G. and others},
  journal={Field Robotics},
  pages={1657--1677},
  year={2023}
}

@inproceedings{fankhauser2018robust,
  title={Robust rough-terrain locomotion with a quadrupedal robot},
  author={Fankhauser, P. and Bjelonic, M. and others},
  booktitle={IEEE International Conference on Robotics and Automation (ICRA)},
  pages={5761--5768},
  year={2018},
  organization={IEEE}
}

@article{Suyoung2023deformter,
  title={Learning quadrupedal locomotion on deformable terrain},
  author={Lee, J. and others},
  journal={Science Robotics},
  volume={8},
  number={74},
  pages={eade2256},
  year={2023}
}

@ARTICLE{8772165,
  title={Dynamic Locomotion on Slippery Ground}, 
  author={Jenelten, F. and Hwangbo, J. and Tresoldi, F. and Bellicoso, C. D. and Hutter, M.},
  journal={IEEE Robotics and Automation Letters (RA-L)}, 
  year={2019},
  volume={4},
  number={4},
  pages={4170--4176},
  doi={10.1109/LRA.2019.2931284}
}

@inproceedings{da2021learning,
  title={Learning a contact-adaptive controller for robust, efficient legged locomotion},
  author={Da, X. and Xie, Z. and others},
  booktitle={Conference on Robot Learning (CoRL)},
  pages={883--894},
  year={2021},
  organization={PMLR}
}

@inproceedings{bjelonic2021whole,
  title={Whole-body mpc and online gait sequence generation for wheeled-legged robots},
  author={Bjelonic, M. and Grandia, R. and Harley, O. and others},
  booktitle={IEEE/RSJ International Conference on Intelligent Robots and Systems (IROS)},
  pages={8388--8395},
  year={2021},
  organization={IEEE}
}

@article{bjelonic2020rolling,
  title={Rolling in the deep--hybrid locomotion for wheeled-legged robots using online trajectory optimization},
  author={Bjelonic, M. and Sankar, P. K. and Bellicoso, C. D. and others},
  journal={IEEE Robotics and Automation Letters (RA-L)},
  volume={5},
  number={2},
  pages={3626--3633},
  year={2020},
  publisher={IEEE}
}

@article{lee2024learning,
  title={Learning robust autonomous navigation and locomotion for wheeled-legged robots},
  author={Lee, J. and Bjelonic, M. and Reske, A. and others},
  journal={Science Robotics},
  volume={9},
  number={89},
  pages={eadi9641},
  year={2024},
  publisher={American Association for the Advancement of Science}
}

@article{bjelonic2022offline,
  title={Offline motion libraries and online MPC for advanced mobility skills},
  author={Bjelonic, M. and Grandia, R. and others},
  journal={The International Journal of Robotics Research (IJRR)},
  volume={41},
  number={9-10},
  pages={903--924},
  year={2022},
  publisher={SAGE Publications Sage UK: London, England}
}

@article{jelavic2023lstp,
  title={Lstp: Long short-term motion planning for legged and legged-wheeled systems},
  author={Jelavic, E. and Qu, K. and Farshidian, F. and Hutter, M.},
  journal={IEEE Transactions on Robotics (TR-O)},
  year={2023},
  publisher={IEEE}
}

@article{zhou2023max,
  title={Max: A Wheeled-Legged Quadruped Robot for Multimodal Agile Locomotion},
  author={Zhou, Q. and Yang, S. and Jiang, X. and others},
  journal={IEEE Transactions on Automation Science and Engineering},
  year={2023},
  publisher={IEEE}
}

@article{geilinger2020computational,
  title={A computational framework for designing skilled legged-wheeled robots},
  author={Geilinger, M. and Winberg, S. and Coros, S.},
  journal={IEEE Robotics and Automation Letters (RA-L)},
  volume={5},
  number={2},
  pages={3674--3681},
  year={2020},
  publisher={IEEE}
}

@inproceedings{bjelonic2018skating,
  title={Skating with a force controlled quadrupedal robot},
  author={Bjelonic, M. and Bellicoso, C. D. and others},
  booktitle={IEEE/RSJ International Conference on Intelligent Robots and Systems (IROS)},
  pages={7555--7561},
  year={2018},
  organization={IEEE}
}

@article{chen2023locomotion,
  title={Locomotion Control of Quadrupedal Robot With Passive Wheels Based on CoI Dynamics on SE (3)},
  author={Chen, J. and Xu, K. and Qin, R. and Ding, X.},
  journal={IEEE Transactions on Industrial Electronics},
  year={2023},
  publisher={IEEE}
}

@inproceedings{tenreiro2006overview,
  title={An overview of legged robots},
  author={Tenreiro Machado, J. A. and Silva, M.},
  booktitle={International Symposium on Mathematical Methods in Engineering},
  pages={1--40},
  year={2006}
}

@inproceedings{valsecchi2023towards,
  title={Towards legged locomotion on steep planetary terrain},
  author={Valsecchi, G. and Weibel, C. and others},
  booktitle={IEEE/RSJ International Conference on Intelligent Robots and Systems (IROS)},
  pages={786--792},
  year={2023},
  organization={IEEE}
}

@inproceedings{xin2017torque,
  title={A torque-controlled humanoid robot riding on a two-wheeled mobile platform},
  author={Xin, S. and You, Y. and Zhou, C. and others},
  booktitle={IEEE/RSJ International Conference on Intelligent Robots and Systems (IROS)},
  pages={1435--1442},
  year={2017},
  organization={IEEE}
}

@inproceedings{gong2019feedback,
  title={Feedback control of a cassie bipedal robot: Walking, standing, and riding a segway},
  author={Gong, Y. and Hartley, R. and Da, X. and others},
  booktitle={American Control Conference (ACC)},
  pages={4559--4566},
  year={2019},
  organization={IEEE}
}

@inproceedings{chen2019feedback,
  title={Feedback control for autonomous riding of hovershoes by a cassie bipedal robot},
  author={Chen, S. and Rogers, J. and others},
  booktitle={IEEE-RAS International Conference on Humanoid Robots},
  pages={1--8},
  year={2019},
  organization={IEEE}
}

@inproceedings{kimura2018riding,
  title={Riding and speed governing for parallel two-wheeled scooter based on sequential online learning control by humanoid robot},
  author={Kimura, K. and Nozawa, S. and others},
  booktitle={IEEE/RSJ International Conference on Intelligent Robots and Systems (IROS)},
  pages={1--9},
  year={2018},
  organization={IEEE}
}

@inproceedings{rajendran2022towards,
  title={Towards humanoids using personal transporters: Learning to ride a segway from humans},
  author={Rajendran, V. and Lin, J. F.-S. and Mombaur, K.},
  booktitle={IEEE RAS/EMBS International Conference for Biomedical Robotics and Biomechatronics},
  pages={01--08},
  year={2022},
  organization={IEEE}
}

@article{margolis2024rapid,
  title={Rapid locomotion via reinforcement learning},
  author={Margolis, G. B. and Yang, G. and Paigwar, K. and others},
  journal={The International Journal of Robotics Research (IJRR)},
  volume={43},
  number={4},
  pages={572--587},
  year={2024},
  publisher={SAGE Publications Sage UK: London, England}
}

@inproceedings{nguyen2004segway,
  title={Segway robotic mobility platform},
  author={Nguyen, H. G. and Morrell, J. and others},
  booktitle={Mobile Robots XVII},
  volume={5609},
  pages={207--220},
  year={2004},
  organization={SPIE}
}

@misc{zapata2024flyboard,
  author={Zapata, F.},
  title={{Flyboard Air}},
  year={2016},
  howpublished={\url{https://www.zapata.com/flyboard-air-by-franky-zapata/}}
}

@misc{rosrobosavvybalance2017,
  author={RoboSavvy Ltd.},
  title={RoboSavvy-Balance},
  howpublished={\url{http://wiki.ros.org/Robots/RoboSavvy-Balance}},
  year={2017}
}

@inproceedings{bouabdallah2007full,
  title={Full control of a quadrotor},
  author={Bouabdallah, S. and Siegwart, R.},
  booktitle={IEEE/RSJ international conference on intelligent robots and systems},
  pages={153--158},
  year={2007},
  organization={IEEE}
}

@inproceedings{zhuang2023robot,
  title={Robot Parkour Learning},
  author={Zhuang, Z. and Fu, Z. and Wang, J. and others},
  booktitle={Conference on Robot Learning (CoRL)},
  year={2023}
}

@inproceedings{gangwani2020learning,
  title={Learning belief representations for imitation learning in pomdps},
  author={Gangwani, T. and Lehman, J. and Liu, Q. and Peng, J.},
  booktitle={Uncertainty in Artificial Intelligence},
  pages={1061--1071},
  year={2020},
  organization={PMLR}
}

@inproceedings{meng2021memory,
  title={Memory-based deep reinforcement learning for pomdps},
  author={Meng, L. and Gorbet, R. and Kuli{\'c}, D.},
  booktitle={IEEE/RSJ International Conference on Intelligent Robots and Systems (IROS)},
  pages={5619--5626},
  year={2021},
  organization={IEEE}
}

@inproceedings{Yu2017si,
  title={Preparing for the Unknown: Learning a Universal Policy with Online System Identification},
  author={Yu, W. and Tan, J. and Liu, C. K. and Turk, G.},
  booktitle={Robotics: Science and Systems},
  year={2017}
}

@inproceedings{kumar2021rma,
  title={Rma: Rapid motor adaptation for legged robots},
  author={Kumar, A. and Fu, Z. and Pathak, D. and Malik, J.},
  booktitle={Robotics: Science and Systems},
  year={2021}
}

@article{lee2020learning,
  title={Learning quadrupedal locomotion over challenging terrain},
  author={Lee, J. and Hwangbo, J. and Wellhausen, L. and others},
  journal={Science Robotics},
  volume={5},
  number={47},
  pages={eabc5986},
  year={2020},
  publisher={American Association for the Advancement of Science}
}

@article{miki2022learning,
  title={Learning robust perceptive locomotion for quadrupedal robots in the wild},
  author={Miki, T. and Lee, J. and Hwangbo, J. and others},
  journal={Science Robotics},
  volume={7},
  number={62},
  pages={eabk2822},
  year={2022},
  publisher={American Association for the Advancement of Science}
}

@article{ji2022concurrent,
  title={Concurrent training of a control policy and a state estimator for dynamic and robust legged locomotion},
  author={Ji, G. and Mun, J. and Kim, H. and Hwangbo, J.},
  journal={IEEE Robotics and Automation Letters (RA-L)},
  volume={7},
  number={2},
  pages={4630--4637},
  year={2022},
  publisher={IEEE}
}

@inproceedings{cheng2024extreme,
  title={Extreme parkour with legged robots},
  author={Cheng, X. and Shi, K. and Agarwal, A. and Pathak, D.},
  booktitle={IEEE International Conference on Robotics and Automation (ICRA)},
  pages={11443--11450},
  year={2024},
  organization={IEEE}
}

@inproceedings{fu2023deep,
  title={Deep whole-body control: learning a unified policy for manipulation and locomotion},
  author={Fu, Z. and Cheng, X. and Pathak, D.},
  booktitle={Conference on Robot Learning (CoRL)},
  pages={138--149},
  year={2023},
  organization={PMLR}
}

@inproceedings{margolis2023walk,
  title={Walk these ways: Tuning robot control for generalization with multiplicity of behavior},
  author={Margolis, G. B. and Agrawal, P.},
  booktitle={Conference on Robot Learning (CoRL)},
  pages={22--31},
  year={2023},
  organization={PMLR}
}

@article{peng2018deepmimic,
  title={Deepmimic: Example-guided deep reinforcement learning of physics-based character skills},
  author={Peng, X. B. and Abbeel, P. and others},
  journal={ACM Transactions on Graphics (TOG)},
  volume={37},
  number={4},
  pages={1--14},
  year={2018},
  publisher={ACM New York, NY, USA}
}

@article{makoviychuk2021isaac,
  title={Isaac gym: High performance gpu-based physics simulation for robot learning},
  author={Makoviychuk, V. and Wawrzyniak, L. and Guo, Y. and others},
  journal={arXiv preprint arXiv:2108.10470},
  year={2021}
}

@article{schulman2017proximal,
  title={Proximal policy optimization algorithms},
  author={Schulman, J. and Wolski, F. and Dhariwal, P. and others},
  journal={arXiv preprint arXiv:1707.06347},
  year={2017}
}

@article{taheri2023study,
  title={A study on quadruped mobile robots},
  author={Taheri, H. and Mozayani, N.},
  journal={Mechanism and Machine Theory},
  volume={190},
  pages={105448},
  year={2023},
  publisher={Elsevier}
}

@article{yang2014spline,
  title={Spline-based RRT path planner for non-holonomic robots},
  author={Yang, K. and Moon, S. and Yoo, S. and others},
  journal={Journal of Intelligent \& Robotic Systems},
  volume={73},
  number={1},
  pages={763--782},
  year={2014},
  publisher={Springer}
}

@article{samuel2016review,
  title={A review of some pure-pursuit based path tracking techniques for control of autonomous vehicle},
  author={Samuel, M. and Hussein, M. and Mohamad, M. B.},
  journal={The International Journal of Computer Applications},
  volume={135},
  number={1},
  pages={35--38},
  year={2016},
  publisher={Foundation of Computer Science}
}

@article{abdulghany2017generalization,
  title={Generalization of parallel axis theorem for rotational inertia},
  author={Abdulghany, A. R.},
  journal={American Journal of Physics},
  volume={85},
  number={10},
  pages={791--795},
  year={2017}
}

@inproceedings{siddhardha2019quadrotor,
  title={Quadrotor hoverboard},
author={Siddhardha, K. and Manathara, J. G.},
  booktitle={Indian Control Conference},
  pages={19--24},
  year={2019},
  organization={IEEE}
}

@misc{hendohover,
  title = {{Hendo Hoverboard}},
  howpublished = {\url{https://hendohover.com/}},
  year = {2015},
  author = {{Arx Pax, LLC}}
}

@misc{hover1hoverboards,
  title = {{Hover-1 Hoverboards}},
  howpublished = {\url{https://www.hover-1.com/collections/hoverboards}},
  year = {2017},
  author = {Frank, D.}
}
    }

\end{document}